\documentclass[a4paper]{article}

\usepackage{INTERSPEECH2018}
\usepackage{amsmath,graphicx,color}
\usepackage{array}
\usepackage{longtable}
\usepackage{cite,amssymb}
\usepackage{subcaption}
\usepackage{authblk}
\usepackage{url}

\title{Transfer Learning for Improving Speech Emotion Classification Accuracy}
\name{Siddique Latif$^{1,3}$, Rajib Rana$^2$, Shahzad Younis$^3$, Junaid Qadir$^1$, Julien Epps$^4$}
\address{
  $^1$Information Technology University (ITU)-Punjab, Pakistan\\
  $^2$University of Southern Queensland, Australia\\
$^3$National University of Sciences and Technology (NUST), Pakistan\\
$^4$The University of New South Wales, Sydney, Australia}
\email{siddique.latif@itu.edu.pk, rajib.rana@usq.edu.au, muhammad.shahzad@seecs.edu.pk, junaid.qadir@itu.edu.pk, j.epps@unsw.edu.au}

\begin{document}

\maketitle
\begin{abstract}
 The majority of existing speech emotion recognition research focuses on automatic emotion detection using training and testing data from same corpus collected under the same conditions. The performance of such systems has been shown to drop significantly in cross-corpus and cross-language scenarios. To address the problem, this paper exploits a transfer learning technique to improve the performance of speech emotion recognition systems that is novel in cross-language and cross-corpus scenarios. Evaluations on five different corpora in three different languages show that Deep Belief Networks (DBNs) offer better accuracy than previous approaches on cross-corpus emotion recognition, relative to a Sparse Autoencoder and SVM baseline system. Results also suggest that using a large number of languages for training and using a small fraction of the target data in training can significantly boost accuracy compared with baseline also for the corpus with limited training examples. 
\end{abstract}
\noindent\textbf{Index Terms}: cross-corpus, speech, emotion recognition, Deep Belief Networks

\section{Introduction}
\label{sec:intro}

In recent years, speech emotion recognition has received increasing interest. Speech emotion recognition focuses on using linguistic and acoustic attributes as input features and machine learning models as classifiers to classify the emotions of the speaker \cite{batliner2011automatic}. These systems achieve promising results when training and testing are performed from the same corpus. However, for real applications, such systems have been demonstrated not to perform well when speech utterances from different languages and different age groups, in quite different conditions, are combined \cite{schuller2012synthesized}. 

At present, various emotional corpora exist, but they are dissimilar in terms of the spoken language, type of emotion (i.e., naturalistic, elicited, or acted) and labelling scheme (i.e., dimensional or categorical) \cite{schuller2010cross}. There are more than 5,000 spoken languages around the world, but only 389 languages account for 94\% of the world's population\footnote{\url{https://www.ethnologue.com/statistics}}. Even for 389 languages, very few adequate resources (speech corpus) are available for language and speech processing research. This means that research in language and speech analysis must confront the problem of data scarcity for many languages. This imbalance, variation,  diversity, and dynamics in speech and language databases means that it is almost impossible to learn a model from a single corpus and then expect it to be effective in practice in general. 
  
In automatic speech emotion recognition, most studies focus on a single corpus at a time, without considering the performance of model in cross-language and cross-corpus scenarios.
However, ever since transfer learning has been applied to cross-domain classification and pattern recognition problems, interest in applying it to cross-corpus emotion recognition has bee growing. Transfer learning focuses on adapting knowledge from available auxiliary resources to transfer this learning to a target domain, where a very few or even no labelled data is available \cite{pan2010survey,lu2015transfer}. 

Deep neural network (DNN) based transfer learning has recently improved image classification by using a very large dataset as source domain and small data as a target domain \cite{sawada2015transfer}. 
Inspired by this success, deep learning based transfer learning has recently been used for speech analysis. However, the existing research has focused on basic DNNs. The impact of using models like Deep Belief Networks (DBNs), which have strong generalisation power and are therefore suitable for cross-corpus emotion recognition, has not been thoroughly explored. A few studies have explored DBNs for speech emotion recognition (e.g., \cite{le2013emotion,rana2016emotion}) and numerous studies focus on DBNs for features extraction \cite{xia2017multi,schmidt2011learning,huang2014research} from speech signal. However, transfer learning using DBNs is very rare. Furthermore, how to maximise the transfer learning performance for cross-corpus/cross-language emotion recognition still needs to be explored further.


In this study, we address the above challenges. We investigate DBNs for transfer learning over five widely-used emotional speech databases. By using the experimental results from various scenarios, we indicated how a large gain in accuracy comparable to baseline can be achieved using transfer learning technique for cross-corpus emotion recognition.



\section{Related Work}
\label{sec:back}

Although cross-language and cross-corpus speech emotion recognition is an interesting problem, relatively few studies have addressed this topic. Existing studies have mostly studied the preliminary feasibility of cross-corpus learning and pointed to the need for further in-depth research. For example, Schuller et al. \cite{schuller2010cross} used six different corpora to analyse cross-corpora emotion recognition using support vector machines (SVM) and highlighted the limitations of current systems for cross-corpus emotion recognition. Eyben et al. \cite{eyben2010cross} used four corpora to evaluate some pilot experiments on cross-corpus emotion recognition while using SVM. They used three datasets for training and a fourth for testing, and showed that the cross-corpus emotion recognition is feasible. To explore the universal cues of emotions across languages, Xia et al. \cite{xiao2016speech} investigated cross-language emotion recognition for Mandarin vs. Western languages (i.e., German, and Danish). The authors focused on gender-specific speech emotion recognition and achieved the classification rates higher than the chance level but less than baseline accuracy.  Albornoz et al. \cite{albornoz2017emotion} developed an ensemble SVM for emotion detection with a focus on emotion recognition in unseen languages.

 Deep learning techniques have been widely used for transfer learning in speech recognition but only basic DNN models have been utilised so far. Lim et al. \cite{lim2016cross} proposed cross-acoustic transfer learning framework by using DNNs. The authors trained a model on a large data of speech and use it for sound event classification. After a series of experiments, the results showed that the cross-acoustic transfer learning can significantly enhance the sound event classification rate. In \cite{richardson2015deep}, authors used a single DNN for speaker and language recognition with a large gain on performance by training the model on speech recognition data. 
 These studies exploited the models that have good learning abilities so that the learned features are transferable to enable model adaptation regarding the target domain.


In this paper, we use Deep Belief Networks (DBNs) for transfer learning  speech emotion. The key reason for employing DBN is its power of generalisation, which is not present in most conventional DNN models \cite{lee2010unsupervised}. Because, the building block of DBNs (i.e., RBMs) are universal approximators and very powerful to approximate any distribution \cite{le2008representational}.
Intuitively, for cross-corpus and cross-language emotion recognition, the generalisation power of a model is crucial. In addition, DBN can learn more powerful and effective discriminative long-range of features \cite{hinton2006reducing} that have been shown to help in speech-related problems \cite{deng2010binary}.


Apart from DNNs, researchers have also used interesting deep architectures for transfer learning. In \cite{gideon2017progressive}, the authors focused on using Progressive Neural Networks to transfer knowledge for three paralinguistic tasks, i.e., emotion, speaker, and gender detection. Progressive Networks are useful for conducting multitasking in a network, however, we focus on a single task of emotion recognition as speaker and gender recognition are not the focus of this paper. Zong et al. \cite{zong2016cross} proposed a domain-adaptive least-squares regression (DaLSR) model for cross-corpus speech emotion recognition. They used three datasets for the evaluations and found that DaLSR can achieved better results than other models like SVM. They did not focus on achieving results higher than the baseline accuracy. Similarly, Deng et al. \cite{deng2013sparse} used sparse autoencoders (AE) for feature transfer learning in speech emotion recognition.  They used six standard databases and a single-layer sparse AE and train this model on class-specific instances from the target domain, then apply this representation to the source domain for reconstruction of those data. This experimental approach improves the performance of the model as compared with independent learning from every source domain.

\section{Experimental Setup}
\label{ES}
\subsection{Speech Databases}
\label{sec:datasets}

\begin{table*}[ht]
\centering
\scriptsize
\caption{Corpora information and the mapping of class labels onto Negative/Positive valence.}
\begin{tabular}{|m{1.2cm}|m{1.1cm}|m{1cm}|m{1.2cm}|m{4.15cm}|m{3.28cm}|m{1.2cm}|}
\hline
\textbf{Corpus}
&\textbf{Language}
&\textbf{Age}
&\textbf{Utterances}
&\textbf{Negative Valance}
&\textbf{Positive Valance}
&\textbf{References}
\\ \hline
\begin{tabular}[c]{@{}l@{}}FAU-AIBO\end{tabular}
&\begin{tabular}[c]{@{}l@{}}German\end{tabular}
&\begin{tabular}[c]{@{}l@{}}Children\end{tabular}
&\begin{tabular}[c]{@{}l@{}}18216\end{tabular}
&\begin{tabular}[c]{@{}l@{}}Angry, Touchy, Emphatic,
Reprimanding\end{tabular}
&\begin{tabular}[c]{@{}l@{}}Motherese, Joyful, Neutral,
Rest\end{tabular}
&\begin{tabular}[c]{@{}l@{}}\cite{schuller2009interspeech}\end{tabular}
\\\hline
\begin{tabular}[c]{@{}l@{}}IEMOCAP\end{tabular}
&\begin{tabular}[c]{@{}l@{}}English\end{tabular}
&\begin{tabular}[c]{@{}l@{}}Adults\end{tabular}
&\begin{tabular}[c]{@{}l@{}}5531\end{tabular}
&\begin{tabular}[c]{@{}l@{}}Angry, Sadness\end{tabular}
&\begin{tabular}[c]{@{}l@{}}Neutral, Happy, Excited\end{tabular}
&\begin{tabular}[c]{@{}l@{}}\cite{busso2008iemocap}\end{tabular}
\\\hline
 \begin{tabular}[c]{@{}l@{}}EMO-DB\end{tabular}
&\begin{tabular}[c]{@{}l@{}}German\end{tabular}
&\begin{tabular}[c]{@{}l@{}}Adults\end{tabular}
&\begin{tabular}[c]{@{}l@{}}494\end{tabular}
&\begin{tabular}[c]{@{}l@{}}Anger, Sadness, Fear, Disgust, Boredom\end{tabular} 
&\begin{tabular}[c]{@{}l@{}}Neutral, Happiness\end{tabular}
&\begin{tabular}[c]{@{}l@{}}\cite{burkhardt2005database}\end{tabular}
\\ \hline
 \begin{tabular}[c]{@{}l@{}}SAVEE\end{tabular}
&\begin{tabular}[c]{@{}l@{}}English\end{tabular}
&\begin{tabular}[c]{@{}l@{}}Adults\end{tabular}
&\begin{tabular}[c]{@{}l@{}}480\end{tabular}
&\begin{tabular}[c]{@{}l@{}}Anger, Sadness, Fear, Disgust \end{tabular}
&\begin{tabular}[c]{@{}l@{}}Neutral, Happiness, Surprise\end{tabular}
&\begin{tabular}[c]{@{}l@{}}\cite{jackson2014surrey}\end{tabular}
\\ \hline 
 \begin{tabular}[c]{@{}l@{}}EMOVO\end{tabular}
&\begin{tabular}[c]{@{}l@{}}Italian\end{tabular}
&\begin{tabular}[c]{@{}l@{}}Adults\end{tabular}
&\begin{tabular}[c]{@{}l@{}}588\end{tabular}
&\begin{tabular}[c]{@{}l@{}}Anger, Sadness, Fear, Disgust\end{tabular}
&\begin{tabular}[c]{@{}l@{}}Neutral, Joy, Surprise\end{tabular}
&\begin{tabular}[c]{@{}l@{}}\cite{costantini2014emovo}\end{tabular}
\\ \hline

\end{tabular}

\centering
\label{table: MAP}
\end{table*}

To investigate the performance of DBN for cross-corpora and cross-language emotion recognition, we selected five publicly available and highly popular corpora which have maximum diversity in languages. These databases are annotated differently, therefore, one of the only consistent ways to investigate transfer learning is by considering the binary positive/negative valence classification problem. We adopt the binary valence mapping per emotion category from \cite{deng2013sparse,eyben2016geneva,schuller2010cross}.  
The names of the datasets used in our experiment and the categorical mappings to binary valence classes are provided in Table \ref{table: MAP}. These databases were chosen to span a variety of languages. 

\subsection{Speech Features}

In this study, we use eGeMAPS feature set, which is a widely used reference feature set for speech emotion recognition studies \cite{gideon2017progressive}. 
The feature set includes Low-Level Descriptor (LLD) features of the speech signal which are described most relevant to emotions by Paralinguistic studies \cite{eyben2016geneva}. The eGeMAPS feature set contains 88 features including frequency, energy, spectral, cepstral, and dynamic information. The overall components are the arithmetic mean and coefficient of variation of 18 LLDs, 6 temporal features, 4 statistics over the unvoiced segments, 8 functionals applied to loudness and pitch, and 26 additional dynamic and cepstral components.

\subsection{Deep Belief Networks}
DBNs are very popular deep architectures that consist of the stack of Restricted Boltzmann Machines (RBMs) to make a powerful probabilistic generative model by using layer-wise training in a greedy manner. RBM is an undirected stochastic neural network consisting of a visible layer, a hidden layer, and a bias unit. Each visible unit of the visible layer is fully connected to hidden units in the hidden layer, and the bias is connected to all the visible units and the hidden units. There is no connection between visible to visible and between hidden to hidden units. RBMs can also be used as classifiers. They are trained on the joint distribution of input data and corresponding labels, then the label is assigned to the new input which has the highest probability under the model. The joint distribution of between visible layer ($v$) and hidden layer ($h$) is given by~\cite{hinton2006fast}: 

\begin{equation}
    P(v,h)= \frac{1}{Z}\exp({-E(v,h)})
\end{equation}

where Z represents the normalisation constant and $E(v, h)$ is an energy
function which is defined as:

\begin{equation}
    E(v,h)=  -\sum_{i=1}^{D}\sum_{j=1}^{k}W_{ij}v_{i}h_{j}-\sum_{i=1}^{D}b_{i}v_{i}-\sum_{j=1}^{k}a_{j}h_{j}
\end{equation}
where $v_{i}$ and $h_{i}$ are the binary states of visible and hidden units. $W_{ij}$ represents the weights of connections between hidden and visible nodes. The conditional probabilities for the visible and hidden units are given by the following equations, where $g$ is the sigmoid function: $g(x)= \frac{1}{1+e^{-x}}$.

\begin{equation}
    P(v_{i}=1|h)= g\big(b_{i}^{v}+ \sum_{j}h_{j}W_{ij}\big)
\end{equation}

\begin{equation}
    P(h_{j}=1|v)= g\big(b_{j}^{h}+ \sum_{i}v_{i}W_{ij}\big)
\end{equation}

An RBM is pre-trained for the maximisation of data log-likelihood $log P (v)$. The stack of generatively pre-trained RBMs constitutes a powerful DBN that can be discriminatively fine-tuned to improve performance. Weight initialisation with pre-training can help the network to avoid poor local minima and give better discriminative results when compared with a neural network initialised by small random weights \cite{erhan2010does}. In this work, we also use layer-by-layer pre-training for DBN. The description of DBNs and their training methodologies can be reviewed in \cite{hinton2002training,hinton2006fast}.

During experimental work, a DBN with three RBM layers was selected, where the first two RBMs have 1000 hidden unit each, and the third RBM have 2000 hidden units with learning rate of $10^{-3}$ and 500 epochs. This configuration was obtained using cross validation experiments on  validation data. 
The other network parameters were chosen by following the setup in \cite{rana2016emotion,keyvanrad2014brief}. 

\section{Results}
\label{sec:experimentation}

In this section, we explore various scenarios for cross-corpus and cross-language speech emotion recognition and conduct experiments to test the scenarios.  

\subsection{Within Corpus Scheme}
\begin{figure}[!ht]
\centering
\centerline{\includegraphics[width=.42\textwidth]{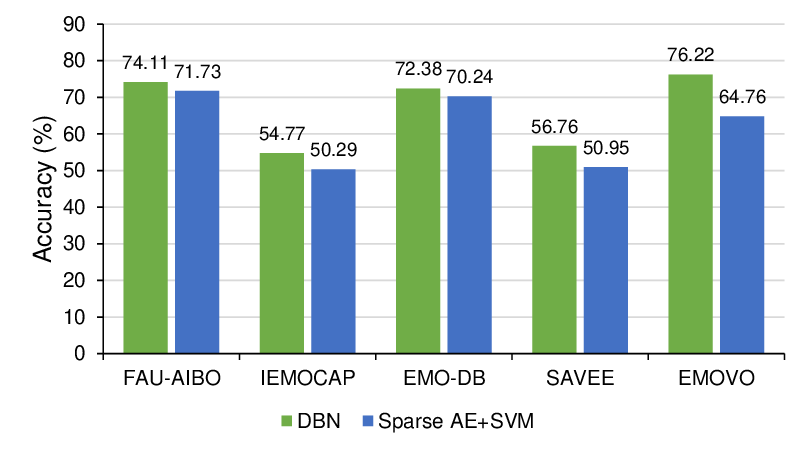}}
\caption{Comparison of baseline accuracy using DBN and sparse AE on different databases.}
\label{fig:base}
\end{figure}

In order to obtain the baseline comparison results, we compare the performance of DBN with a popular approach of using sparse autoencoder (AE) with SVM for feature transfer learning in speech emotion recognition\cite{deng2013sparse}. This preliminary experiment enables us to set maximum achievable baseline accuracy when both systems are trained and tested using the data of same corpus. 
For baseline experiments, 75\% of randomly selected data is used for training and remaining 25\% unseen data is used for testing. 
Figure \ref{fig:base} shows the comparison results, where DBN  outperforms sparse AE for all databases. 


\subsection{Language Tests}
\begin{figure*}[!ht]%
\centering
\begin{subfigure}{0.4\linewidth}
\includegraphics[width=\linewidth]{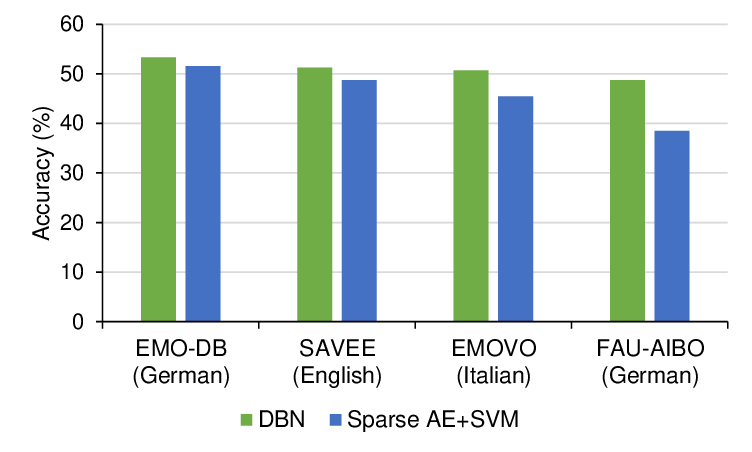}%
\captionsetup{justification=centering}
\caption{}%
\label{Lang1}%
\end{subfigure}
\begin{subfigure}{0.4\linewidth}
\includegraphics[width=\linewidth]{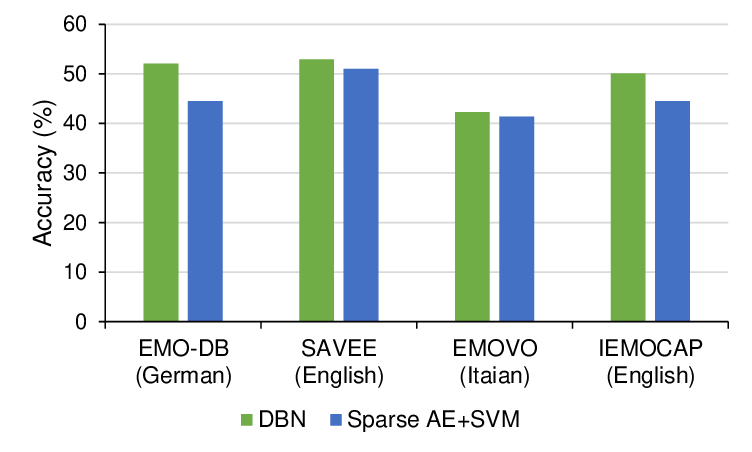}%
\captionsetup{justification=centering}
\caption{} %
\label{lang2}%
\end{subfigure}%
\caption{Comparison of language tests using DBN and sparse AE. Figure \ref{Lang1} represents the recognition rate using IEMOCAP (English) for training and other databases for testing whereas \ref{lang2} shows the recognition rate using FAU-AIBO (German) for training and other databases for testing.}
\label{fig:Lang}
\end{figure*}

In this experiment, we use one language dataset for training and the remaining datasets for testing. For brevity, we just use FAU-AIBO (German) and IEMOCAP (English) datasets for training. 
In order to evaluate the model on IEMOCAP, we used two sessions out of five with two-fold cross validation because overall data is large. The other databases are small comparative to IEMOCAP, therefore, we used them completely. Figure \ref{fig:Lang} shows the recognition rate achieved in these experiments and its comparison with previous techniques using sparse autoencoder and SVM (sparse AE+SVM) for cross-corpus transfer learning. When the IEMOCAP database was used for training the DBN, we performed pairwise testing using OHM and MONT separately for FAU-AIBO. It can be noted from Figure \ref{fig:Lang} that DBN outperforms sparse AE for all scenarios. Beyond this point, the accuracy of sparse AE is not given, as we observe that DBNs consistently outperform sparse AE.

\begin{figure*}[!ht]%
\centering
\begin{subfigure}{0.4\linewidth}
\includegraphics[width=\linewidth]{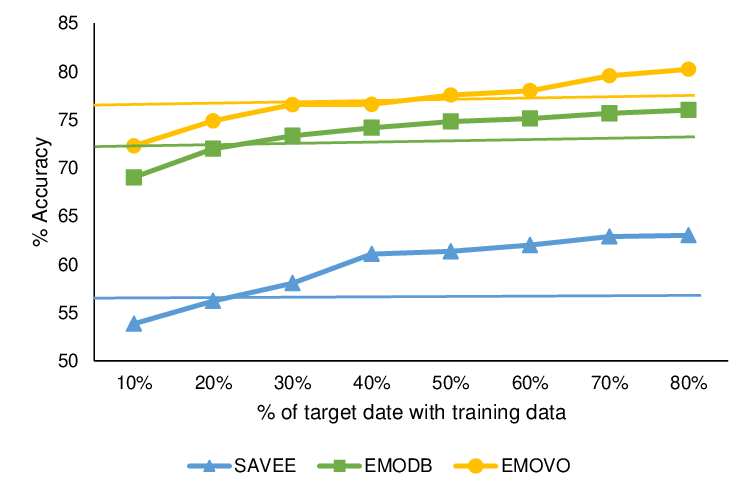}%
\captionsetup{justification=centering}
\caption{}%
\label{Per1}%
\end{subfigure}
\begin{subfigure}{0.4\linewidth}
\includegraphics[width=\linewidth]{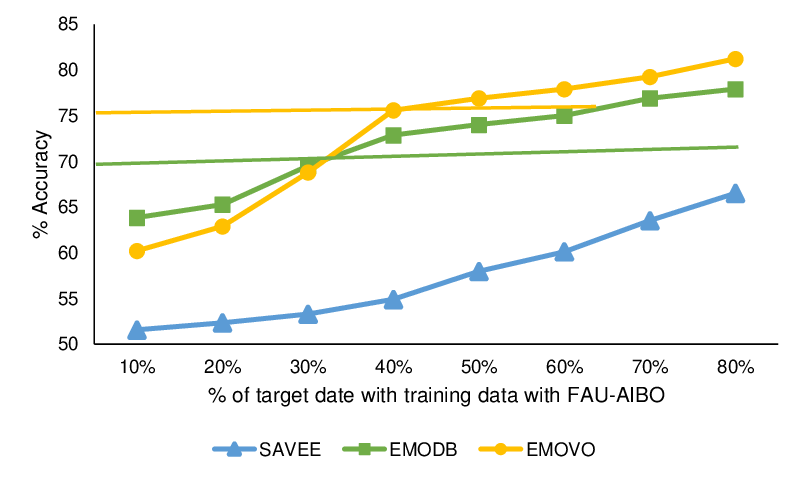}%
\captionsetup{justification=centering}
\caption{} %
\label{Per2}%
\end{subfigure}%
\caption{Impact of using a percentage of target date with training data. Where \ref{Per1} shows the training with IEMOCAP and \ref{Per2} is when training is performed using FAU-AIBO.}
\label{Perentage}
\end{figure*}

\subsection{Percentage of Target Data}

In this experiment, we vary the percentage (10\% to 80\%) of the target dataset for the training of the model. The training was performed using IEMOCAP and FAU-AIBO separately and EMOVO, EMO-DB and SAVEE were used for testing. The results are shown in Figure \ref{Perentage}. The straight horizontal lines in the figure show the baseline recognition rate for the respective corpora. These results show that the recognition rate significantly improves (than baseline) by including target domain data with the training data.

\subsection{Multi-language Training}

In this experiment, we use multiple languages jointly for training to observe whether this improves the performance of using languages individually for training. We use both FAU-AIBO and IEMOCAP for training and remaining for testing. We also evaluate the model within the corpora. For IEMOCAP, we used three sessions (plus FAU-AIBO) for training  and testing was performed using the remaining two sessions with two-fold cross validation. Similarly, for FAU-AIBO, a two-fold cross-validation was used, i.e., training on OHM (plus IEMOCAP) and evaluating on MONT and the inverse.

\begin{figure}[!ht]
\centering
\captionsetup{justification=centering}
\centerline{\includegraphics[width=.42\textwidth]{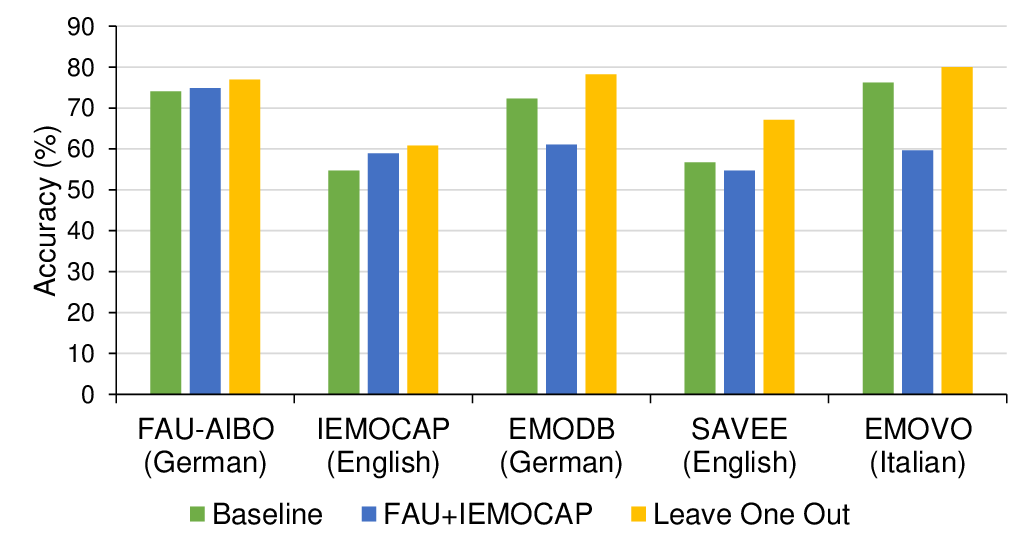}}
\caption{Comparison of baseline results and transfer learning using FAU-AIBO+IEMOCAP and leave-one-language-out scheme.}
\label{fig:G2}
\end{figure}

Further, we also performed training using a leave-one-data-out scheme. For FAU-AIBO, we have performed evaluation by using OHM and MONT independently taking the average results. In the case of IEMOCAP, we used two sessions (with two-fold cross validation) to evaluate the model. This performs better than baseline and two-language training as shown in Figure \ref{fig:G2}. 

\section{Discussion}
\label{Sec: Insigts}
From the experiments, Leave-one-Out seems to be standing out in-terms of obtaining the highest accuracy. This essentially means that training the model using a large range of languages would help learn many intrinsic features from each languages, which can essentially help to achieve high accuracy in an unknown language - even higher than when the same language is used for training and testing (baseline). The performance of the Leave-one-out (see Figure~\ref{fig:G2}) on EMOVO database is a prime example of this. Both German and English languages have two datasets each, i.e., in a Leave-one-Out scheme there will be at least one of these language in the training set. But for EMOVO there will be a situation that emotions in the Italian language are predicted simply based on emotions in German and English language. 

Another interesting aspect we learned from the experiments that including a fraction of the target data into training can help improve the performance and help achieve better results than baseline. 
Based on our experiments, augmenting other databases with around 20\% of data (around 90 utterances in case of EMO-DB) from the target database can help achieve better than the baseline accuracy. However, this is worse while using FAU-AIBO for training. Interestingly, IEMOCAP performs well on EMO-DB that is in the German language as compared to FAU-AIBO that is also in German. We note that FAU-AIBO consists of children speech whereas EMO-DB database contains adult speech.
 

The performance of DBN in the language test results in Figure~\ref{fig:Lang} using both IEMOCAP (English) and FAU-AIBO (German) on target datasets is poor than the baseline. The drop in accuracy is not only for the target dataset with a different language but also for target data having similar language. From this experiment, we learned that the different studio conditions, age and language differences, and type of emotional corpus cause drop in the performance of the model. This problem can be addressed by previous two findings, i.e., either by training the model with the data of multiple languages or by including a small portion of data target domain with training data. 



\section{Conclusions}
\label{sec: con}

In this paper, we investigated the performance of DBNs for transfer learning based cross-corpus and cross-language speech emotion recognition. In order to evaluate the feature transference across different corpora, we performed comprehensive experiments and found that DBNs outperformed sparse autoencoders due to its increased feature learning abilities. Also,  DBNs can learn from many training languages and improve the baseline accuracy even also when a small fraction of target data is included in the model while training it with a single corpus. For practical applications, these findings would be very helpful to build a robust speech emotion recognition system using data from multiple languages. Also, this would be equally useful for emotion recognition in languages with very limited or no datasets.



\end{document}